\pdfoutput=1

\documentclass[11pt]{article}

\usepackage[final]{coling}

\usepackage{times}
\usepackage{latexsym}

\usepackage[T1]{fontenc}

\usepackage[utf8]{inputenc}

\usepackage{microtype}

\usepackage{inconsolata}

\usepackage{graphicx}

\usepackage[utf8]{inputenc} 
\usepackage[T1]{fontenc}    
\usepackage{hyperref}       
\usepackage{booktabs}       
\usepackage{amsfonts}       
\usepackage{nicefrac}       
\usepackage{microtype}      
\usepackage{xcolor}         
\usepackage{graphicx}
\usepackage{amsmath}
\usepackage{subcaption} 
\usepackage{multirow}
\usepackage{makecell}
\usepackage{array}    

\makeatletter
\def\NAT@force@numbers{\NAT@force@numbers}
\makeatother

\usepackage{multirow}
\usepackage{geometry}
\usepackage{arydshln} 
\usepackage{amsmath,bm}
\usepackage{ulem}
\usepackage{pifont}
\usepackage{xcolor}
\usepackage{soul}
\usepackage{soulpos}

\title{\textbf{\textsc{ConTrans}}: Weak-to-Strong Alignment Engineering via Concept Transplantation}

\author{
  Weilong Dong\textsuperscript{1} \quad Xinwei Wu\textsuperscript{1} \quad Renren Jin\textsuperscript{1} \quad Shaoyang Xu\textsuperscript{2} \quad Deyi Xiong\textsuperscript{1,}\textsuperscript{2}\thanks{Corresponding author.}\\
  \textsuperscript{1}College of Intelligence and Computing, Tianjin University \\
  \textsuperscript{2}School of New Media and Communication, Tianjin University \\
  \texttt{\{willowd, wuxw2021, rrjin, syxu, dyxiong\}@tju.edu.cn} 
}

\begin{document}
\maketitle
\begin{abstract}
Ensuring large language models (LLM) behave consistently with human goals, values, and intentions is crucial for their safety but yet computationally expensive. 
To reduce the computational cost of alignment training of LLMs, especially for those with a huge number of parameters, and to reutilize learned value alignment, we propose \textbf{\textsc{ConTrans}}, a novel framework that enables weak-to-strong alignment transfer via concept transplantation.  
From the perspective of representation engineering, \textbf{\textsc{ConTrans}} refines concept vectors in value alignment from a source LLM (usually a weak yet aligned LLM). The refined concept vectors are then reformulated to adapt to the target LLM (usually a strong yet unaligned base LLM) via affine transformation. In the third step, \textbf{\textsc{ConTrans}} transplants the reformulated concept vectors into the residual stream of the target LLM. 
Experiments demonstrate the successful transplantation of a wide range of aligned concepts from 7B models to 13B and 70B models across multiple LLMs and LLM families. Remarkably, \textbf{\textsc{ConTrans}} even surpasses instruction-tuned models in terms of truthfulness. 
Experiment results validate the effectiveness of both inter-LLM-family and intra-LLM-family concept transplantation. 
Our work successfully demonstrates an alternative way to achieve weak-to-strong alignment generalization and control. 
The code is available at \href{https://github.com/willowdong/ConTrans}{github.com/willowdong/ConTrans}.

\textcolor{red}{Warning: This paper contains content that may be offensive or harmful.}
\end{abstract}

\section{Introduction}
Large language models are trained on a huge amount of data, which allows them to develop strong capabilities  in a wide array of tasks \citep{brown2020language,touvron2023llama}. However, the training objectives of LLMs at the pre-training stage usually do not align with human goals and values, causing pre-trained models to potentially yield harmful outputs, e.g., biased content or disinformation \citep{perez2022red,jiang2024automated,huang2024cbbq,guo2023evaluatinglargelanguagemodels}. Consequently, ensuring the alignment of LLMs behaviors and their decision process with the goals and values of humans is crucial for the development of safe and trustworthy AI systems.

Various methods have been proposed to align LLMs with human values and preferences, including supervised fine-tuning (SFT) \citep{wei2021finetuned,sanh2022multitask}, reinforcement learning from human feedback (RLHF) \citep{ouyang2022training}, and direct preference optimization (DPO) \citep{rafailov2024direct}. Despite these efforts, significant challenges persist \citep{shen2023large}. First, current alignment methods usually require high-quality and diverse human preference data \citep{wang-etal-2023-self-instruct,zhoustar}, the curation of which is often labor-intensive and time-consuming. Second, the substantial compute required by alignment training is usually not affordable for those with limited computational resources. Third limitations, such as lack of transparency, and training instability \citep{zheng2023secrets} remain with current alignment approaches.

\citet{burns2023weak} propose to utilize weaker models to supervise the training of stronger models, which partially addresses the challenge of sourcing high-quality human-annotated data. The stronger model, trained using labels synthesized by the weaker model, can outperform the weaker model, although it typically falls short of the performance achieved through traditional supervised fine-tuning. Building on this foundation, subsequent studies \citep{li2024superfiltering, chen2024self} have leveraged weak models to filter or generate data for supervised fine-tuning and alignment preference tasks, utilizing the capabilities of smaller models to guide the alignment of larger models. 
All these approaches learn to align stronger models with the data generated by weaker models in an external way as alignment supervision signals are available in the external data.

Distinct from the aforementioned methods, our key interest is to investigate whether weak-to-strong alignment supervision can be achieved internally within the latent feature space of LLMs, from the perspective of representation engineering \citep{zou2023representation}. 
Existing representation engineering methods \citep{li2024inference,wu2024reft} utilize a limited set of positive and negative examples to extract a concept vector, subsequently enhancing the model's preference for the corresponding concept. 
Typically, these methods intervene in the outputs of a model using the concepts that are already embedded in the model. However, if the target model lacks a clear concept, effective intervention becomes unfeasible. Additionally, direct representation engineering across models with different model sizes or from different model families is also impractical as the hidden feature spaces are structurally and dimensionally different.

To overcome these challenges, we propose a novel representation engineering framework, concept transplantation (\textbf{\textsc{ConTrans}}), which facilitates weak-to-strong alignment engineering with three essential components. First, by gathering a small set of positive and negative examples that are semantically related to a given concept, \textbf{\textsc{ConTrans}} refines a semantic vector for the given concept from a source LLM. The refined concept vector is reformulated to be consistent with the feature space of the target LLM via affine transformation. In the last step, the reshaped concept vector is transplanted into the residual stream of the target model, to effectively control the  target LLM output preferences related to the given concept.

Our main contributions are summarized as follows:
\begin{itemize}
    \item We verify that different models possess shared concept features and explore the roles of pre-training and alignment training in the formation and expression of these concepts.
    
    \item We propose \textbf{\textsc{ConTrans}}, a novel framework that enables weak-to-strong alignment engineering from the internal feature space of LLMs via concept transplantation,  which requires no additional training and can effectively transplant a concept using only a few hundred positive and negative examples.
    
    
    \item Our experiments demonstrate the effectiveness of the proposed framework with more than ten LLMs across multiple concepts. 
\end{itemize}

\begin{figure*}[t]
    \vspace{-0.4cm}
    \includegraphics[width=\textwidth]{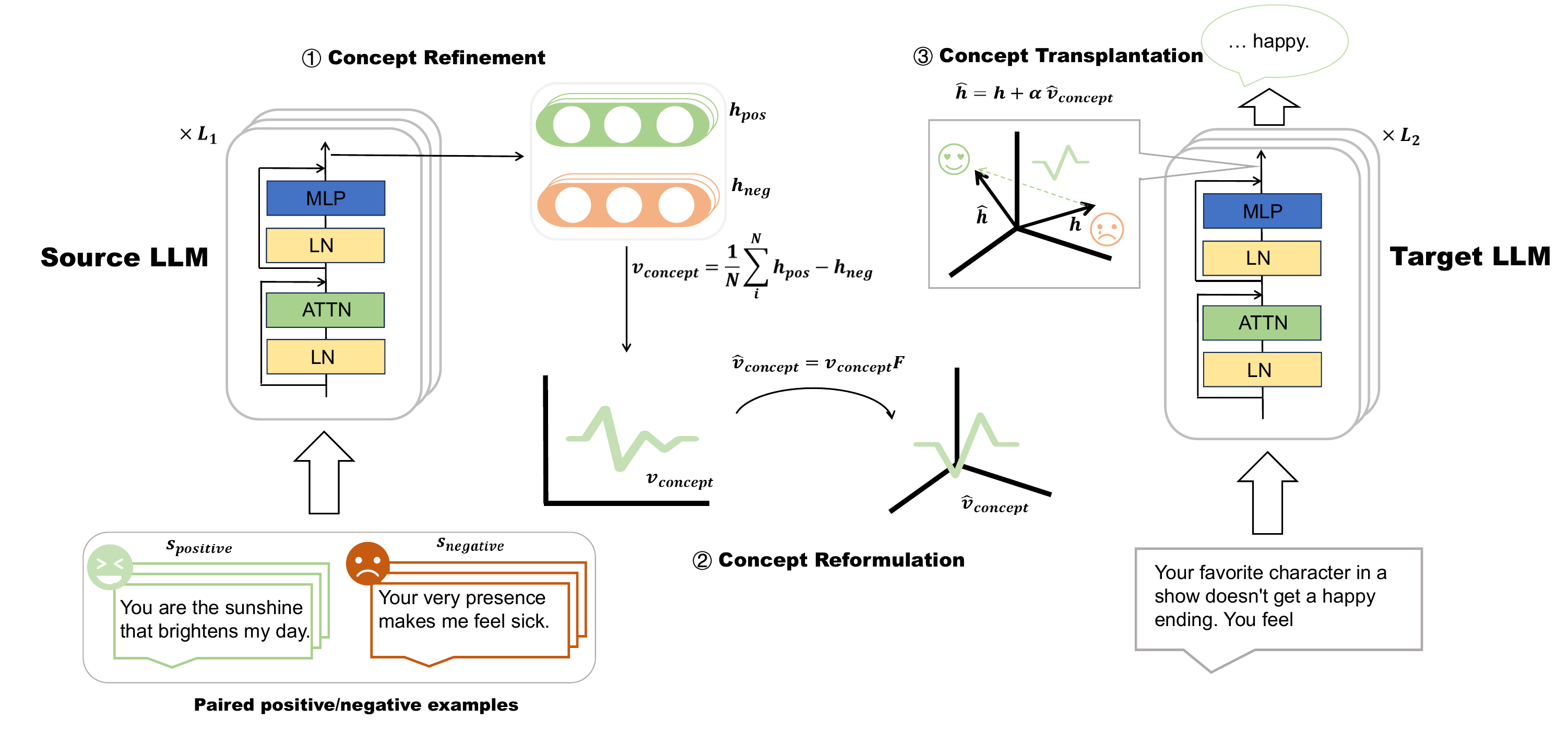}
    \caption{The diagram of \textbf{\textsc{ConTrans}} that consists of three essential modules: \ding{192} concept refinement refining and extracting a vector for a given concept with a set of concept-related positive/negative examples from the source LLM $\mathcal{M}^{\text{src}}$ \ding{193} concept reformulation reshaping and adapting the refined concept vector into the feature space of the target LLM $\mathcal{M}^{\text{tgt}}$ through affine transformation and \ding{194} concept transplantation transplanting the reformulated concept vector into the residual stream of the target LLM to control the outputs of the target LLM related to the given concept.}
    \label{fig:method}
    \vspace{-0.1cm}
\end{figure*}

\section{Related Work}
\label{sec:relatedwork}

\paragraph{Representation Engineering} Representation engineering encompasses a set of techniques that manipulate model outputs by directly modifying the activation values within a model, without altering its parameters. The hidden states of a model encapsulate a wealth of information that are not explicitly manifested in the model's outputs \citep{burns2022discovering,azaria2023internal}. Prior research has focused on targeted interventions for specific concepts such as sentiment \citep{turner2023activation}, truthfulness \citep{li2024inference, zhang2024truthx} and multilingual human values \citep{xu-etal-2024-exploring-multilingual}. 
Other studies aim to enhance intervention methods or analyze the underlying mechanisms of models; for example, \citet{zou2023representation} employ Principal Component Analysis (PCA) to extract and identify concept directions, whereas \citet{wu-etal-2024-mitigating-privacy} reduce privacy leakage in LLMs. 
Distinct from these methods, our work concentrates on cross-model representation engineering instead of representation engineering on the same model, which specifically transplants concepts from small models into the inner space of large models. 

\paragraph{Weak-to-Strong Supervision of LLM} The advent of increasingly powerful AI models poses substantial challenges for alignment. 
The concept of weak-to-strong supervision seeks to utilize weak models to guide the alignment training of strong models. 
\citet{burns2023weak} advocate for using weak models to provide labels to supervise the training of strong models, focusing primarily on enhancing generalization of capabilities. 
\citet{liu2024tuning} and \citet{mitchell2023emulator} suggest intervening in the decoding results of strong models by leveraging the logit differences between aligned and unaligned models. 
\citet{zheng2024weak} deploy a parameter interpolation method to transition from unaligned to aligned models. 
Additionally, \citet{li2024superfiltering}, \citet{chen2024self}, and \citet{ji2024aligner} utilize weak models to filter or generate training data, iteratively enhancing model alignment.
Distinct from these approaches, our work is the first to attempt to perform weak-to-strong alignment internally on the hidden space of LLMs. Previous efforts primarily focus on externally training strong models or treating LLMs as black boxes without delving into the hidden feature space.
Additionally, as \textbf{\textsc{ConTrans}} solely requires the refinement and transplantation of a single concept vector, it does not significantly bring extra cost while previous external weak-to-strong supervision still requires alignment training.

\section{Methodology}
\label{sec:method}

Empirical evidences from both interpretability \citep{tigges2023linear,hernandez2023linearity,park2023linear,nanda2023emergent} and word embeddings \citep{mikolov2013efficient,mikolov2013distributed} support that deep neural networks trained on textual data encapsulate conceptual features.
Recent studies in representation engineering \citep{zou2023representation, li2024inference} have demonstrated that controlled generation in LLMs can be achieved through the manipulation of these representations.  
Building on these empirical evidences and findings, our approach is driven by the hypothesis that concepts are encoded within the feature space of deep neural networks. We hypothesize the existence of population-level representations of concepts within the feature space of LLMs, which are consistent across LLMs of varying sizes and even across LLMs from different model families. By blending the representations in the feature space of the target model with a concept vector from the source model, we can manipulate the polarity along specific conceptual directions in the target model, thereby influencing the expression of these concepts in the generated text.
With this assumption, we propose \textbf{\textsc{ConTrans}} to engineer alignment-related concepts in the target model (strong) with conceptual representations from the source model (weak), providing an internal, transparent and cost-efficient approach to weak-to-strong supervision. 
\textbf{\textsc{ConTrans}} is composed of three essential steps: concept refinement from the source model, concept reformulation and concept transplantation into the target model, illustrated in Figure \ref{fig:method}.

Consider a language model \(\mathcal{M}\) equipped with \(L\) transformer blocks. After tokenization, an input to \(\mathcal{M}\) with $t$ tokens is \(s = (s_{(1)}, s_{(2)}, \ldots, s_{(t)})\). The hidden state of the \(k\)-th layer is hence denoted as \(\bm{h}^k = (\bm{h}_{(1)}^k, \bm{h}_{(2)}^k, \ldots, \bm{h}_{(t)}^k)\), where \(\bm{h}^k \in \mathbb{R}^{t \times d}\) and \(d\) is the feature dimension of \(\mathcal{M}\). The hidden state \(\bm{h}^k\) comprises the residual stream from the preceding layer combined with the output of the multi-layer perceptron (MLP) sublayer in the \(k\)-th layer, which is formulated as \(\bm{h}^k = \bm{h}^{k-1} + \textbf{MLP}(\textbf{ATTN}(\bm{h}^{k-1}))\). \citet{elhage2021mathematical} consider the operations within transformer blocks as interactions within the residual stream, akin to reading and writing processes. Correspondingly, our method can be interpreted as introducing an additional concept vector to the residual stream, thereby adjusting the feature polarity in the targeted direction.

\subsection{Concept Refinement} 
Several methods could be used to refine alignment-related concept representations, including mean difference \citep{turner2023activation}, linear probing \citep{li2024inference}, distributed alignment search (DAS) \citep{geiger2024finding}, and principal component analysis \citep{zou2023representation}. We employ the mean difference method for concept refinement, as it is the most straightforward yet effective approach. The more complex or task-specific concept refinement methods mentioned above are all built upon the mean difference method. We posit that the efficacy of this method can be generalized to more sophisticated approaches.

Given a source language model \(\mathcal{M}^{\text{src}}\) and a target language model \(\mathcal{M}^{\text{tgt}}\), we refine the vector representation of a specific concept from \(\mathcal{M}^{\text{src}}\). Let \(\bm{v}_{\text{concept}}\)  denote the concept vector, (e.g., \(\bm{v}_{\text{happiness}}\) for the concept of happiness).
In order to refine \(\bm{v}_{\text{concept}}\) in the way of representation engineering, it is necessary to use a set of positive and negative textual examples related to the concept. These examples are fed into \(\mathcal{M}^{\text{src}}\) in a forward pass. The hidden state of the last token of the input example at each layer is cached. For a positive example \(s_{\text{positive}}\), the hidden state of the last token is denoted as \(\bm{h}_{\text{pos}}^k\), where \(k \in [1, L]\). 

The concept vector is then refined as the mean difference (direction) between \(\bm{h}_{\text{pos}}\) and \(\bm{h}_{\text{neg}}\):

\begin{equation}
\bm{v}_{\text{concept}}^k = \frac{1}{N} \sum_{i=1}^{N} (\bm{h}_{\text{pos}_{(i)}}^k - \bm{h}_{\text{neg}_{(i)}}^k)
\end{equation}

where \(N\) represents the number of positive/negative example pairs. Given that \(s_{\text{positive}}\) and \(s_{\text{negative}}\) are sentences usually with similar syntax but opposite polarities, the mean difference in their features effectively eliminates lower-level linguistic features while preserving the concept-related feature directions. 

\subsection{Concept Reformulation} The dimensions of $\mathcal{M}^{\text{src}}$ and $\mathcal{M}^{\text{tgt}}$ may differ, which complicates the transplantation of the concept vector refined from \(\mathcal{M}^{\text{src}}\) into \(\mathcal{M}^{\text{tgt}}\). 
To address this issue, we reformulate the refined concept representation by projecting it into a space of different dimension through affine transformation. 
Specifically, the transformation is represented as $\hat{\bm{v}} = \bm{v}\bm{\mathcal{F}},\bm{\mathcal{F}}\in\mathbb{R}^{d_{1}\times d_{2}}$. Here, $d_1$ and $d_2$ respectively denote the hidden dimensions of $\mathcal{M}^{\text{src}}$ and $\mathcal{M}^{\text{tgt}}$. Since the concept may be encoded in a shared low-rank feature space, the concept vector of a weak model can be viewed as a low-rank approximation of the concept vectors of a strong model.
Consequently, the new concept vector can be computed using the equation: $\hat{\bm{v}}_{\text{concept}}^k=\bm{v}_{\text{concept}}^k\bm{\mathcal{F}}$.

To learn the affine transformation $\bm{\mathcal{F}}$, we gather the hidden states $\bm{h}_{\text{src}}$ and $\bm{h}_{\text{tgt}}$ from $\mathcal{M}^{\text{\text{src}}}$ and $\mathcal{M}^{\text{\text{tgt}}}$ using the same input text. 
The objective is to solve for $\bm{\mathcal{F}}$ by minimizing the squared error $\hat{\bm{\mathcal{F}}} = \arg\min_{\bm{\mathcal{F}}} ||\bm{h}_{\text{src}}\bm{\mathcal{F}} - \bm{h}_{\text{tgt}}||^2$. 
The analytical solution is obtained as follows:
\begin{equation}
    \hat{\bm{\mathcal{F}}}=\bm{V} \bm{\Sigma}^{-1} \bm{U}^T \bm{h}_{\text{tgt}}
\end{equation}
 
where the singular value decomposition of $\bm{h}_{\text{src}}$ is $\bm{h}_{\text{src}}=\bm{U}\bm{\Sigma}\bm{V}^{\text{T}}$.
Further details regarding the training of the affine transformation are provided in Appendix \ref{app:affine_train}.

\subsection{Concept Transplantation} Once the concept vector $\bm{v}_{\text{concept}}$ is projected into the feature space of $\mathcal{M}^{\text{tgt}}$,  it can either suppress or enhance the polarity of $\mathcal{M}^{\text{tgt}}$ along the corresponding concept direction. 
We augment the output hidden states of the transformer layers with $\hat{\bm{v}}_{\text{concept}}$. Analogous to the role of a transformer layer, the refined and reformulated concept vector can be considered as introducing new information into the residual stream, serving as an offset in the concept direction of the target model. 
Thus, the hidden states of $k$-th layer will be:
\begin{equation}
    \bm{h}^k=\bm{h}^{k-1}+\textbf{MLP}(\textbf{ATTN}(\bm{h}^{k-1}))+\alpha\hat{\bm{v}}_{\text{concept}}^{k}
\end{equation}
where $\alpha$ is the hyperparameter that controls the strength of steering manipulation. For simplicity, the superscript \(k\) is omitted by default.

\begin{figure*}[t]
    \centering
    \resizebox{0.83\textwidth}{!}{
    \begin{subfigure}{0.5\textwidth}
        \centering
        \includegraphics[width=\linewidth]{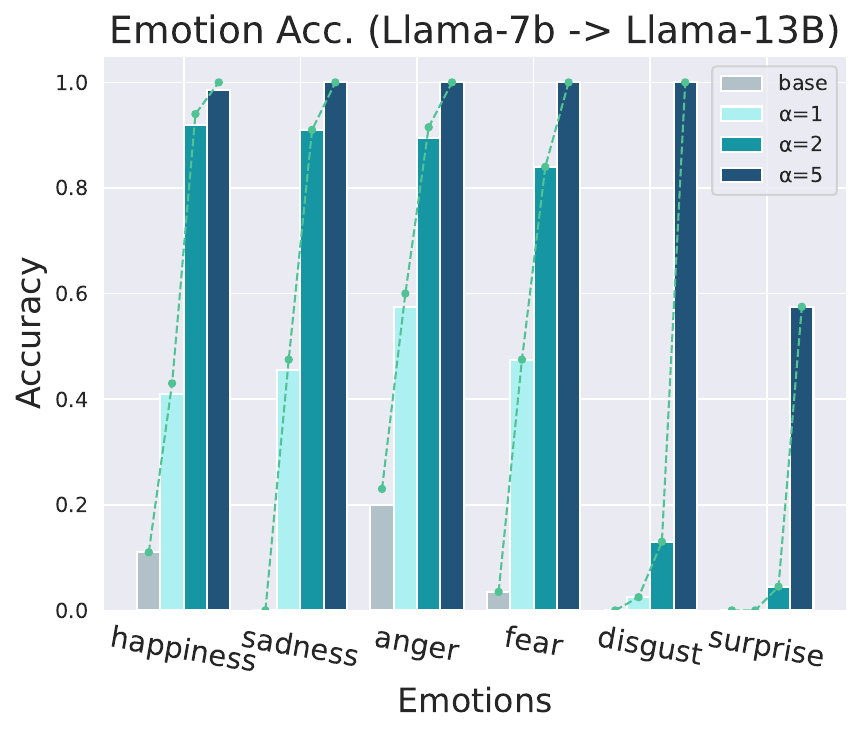}
        \caption{}
        \label{fig:emo_acc}
    \end{subfigure}
    \hfill
    \begin{subfigure}{0.43\textwidth}
        \centering
        \includegraphics[width=\linewidth]{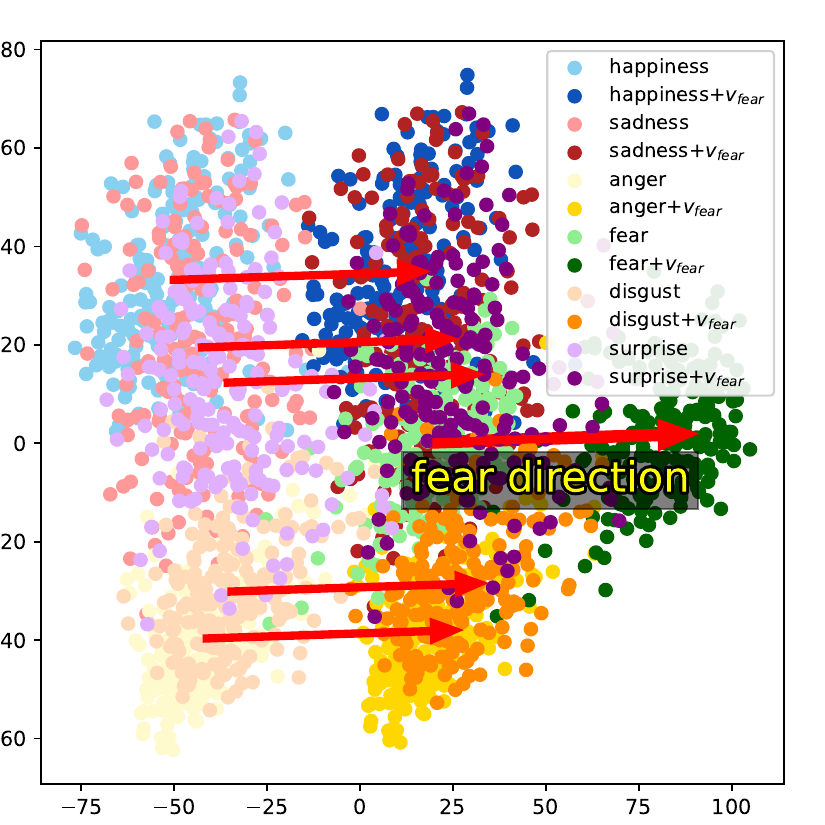}
        \caption{}
        \label{fig:emo_pca}
    \end{subfigure}
    }
    \caption{(a) Emotion prediction accuracy on negative scenarios for each emotion.
    The bar denotes \textbf{Token Acc.}, while the dashed line depicts \textbf{Logit Acc.} (b) The PCA visualization of LLaMA-13B hidden states intervened by \(\bm{v}_{\text{fear}}\) from LLaMA-7B.}
\end{figure*}

\section{Can Concepts Be Transplanted Across Different Models?}
\label{exp:emotion}

To verify the presence of shared concepts between models and to demonstrate that \textbf{\textsc{ConTrans}} can achieve cross-model concept transplantation, 
we conducted experiments and visualizations on the basic concept emotions, which are fundamental concepts in both human cognition and language models. 

\subsection{Intervening in the Emotion of Strong Models}

\paragraph{Setup and Evaluation Metrics} 

We utilized the \textbf{LLaMA} series \citep{touvron2023llama} of models for our experiments, specifically employing the emotion vectors from LLaMA-7B to intervene in the outputs of both LLaMA-13B and LLaMA-65B models. 
We investigated a six-category emotion model—happiness, sadness, anger, fear, surprise, and disgust—and conducted experiments on the emotion dataset introduced by \citet{zou2023representation}. 
Each sentence describes a scenario related to its respective emotion without explicitly incorporating emotion sensitive words. 
To refine a concept vector for each emotion category, we composed positive texts from scenarios that describe the target emotion, while negative texts are derived from scenarios randomly sampled from the other five emotions. 
To better illustrate the impact of concept transplantation, we assessed the model's performance on predicting these negative scenarios. 
Two metrics are employed to measure the prediction accuracy. The first metric, which we term \textbf{Token Acc}, checks if the first token generated by $\mathcal{M}^{\text{\text{tgt}}}$ correctly matches the target emotion token. The second metric, referred to as \textbf{Logits Acc}, measures whether the token with the highest logits among the six emotion tokens is the correct one. 

Results are illustrated in Figure \ref{fig:emo_acc} and Figure \ref{fig:emo_acc_65b}. Given that the negative samples do not depict the target emotion, the baseline accuracy essentially reflects the rate of false positives, which is notably low. 
According to these results, it is evident that $v_{\text{emotion}}$ from LlaMA-7B can effectively influence the emotional response of larger models. Notably, the emotion prediction accuracy of the target model increases with the steering manipulation strength $\alpha$. 
Examples of the controlled generations are presented in Table \ref{tab:case_study}.

\begin{table*}[t]
\centering
\small
\resizebox{0.8\textwidth}{!}{
\begin{tabular}{c|ccc|ccc}
\hline
\multirow{2}{*}{}                              
& \multicolumn{3}{c|}{13B}                            
& \multicolumn{3}{c}{70B}       \\ 
\cline{2-7} 
           & LLaMA 2          & Code LLaMA       & LLaMA           
           & LLaMA 2          & Code LLaMA       & LLaMA                      \\ \hline
Base Model                                
& 17.9\%          & 17.3\%          & 19.0\%          
& 22.1\%          & 19.6\%          & \multicolumn{1}{l}{17.4\%} \\ 
\hline
Align-Training                                
& \textbf{36.8\%}  &\underline{32.9\%} & \textbf{36.7\%} 
& 30.2\%   & 23.7\%          & /      \\ 
Self-Align 
& 35.9\%          & \textbf{33.2\%}          & 23.3\%          
& 28.0\%          & \underline{26.2\%}          & /                          \\ 
Inst-Align                                
& 15.9\%          & 18.5\%          & 17.3\%          
& 19.0\%          & 17.6\%          & 17.7\% \\ 
EFT/proxy-tune                                
& 30.6\%          & 26.1\%          & 30.6\%          
& \underline{31.8\%}          & 25.7\%          & \underline{30.5\%} \\  
\hline
\textbf{\textsc{ConTrans}}
& \underline{36.5\%}          & \underline{32.9\%}          & \underline{30.8\%}     & \textbf{33.9\%} & \textbf{33.4\%} & \textbf{31.8\%}            \\
\hline
\end{tabular}
}

\vspace{1mm}
\caption{
TruthfulQA results for 13B and 70B base models. 
Although the largest size of LLaMA is 65B, we denote it as 70B for notational simplicity.
} 
\label{tab:tqa_main}
\end{table*}

\subsection{Visualization of Concept Intervenation}

We utilized \(\bm{v}_{\text{fear}}\) from LLaMA-7B to intervene in the six emotion hidden states of LLaMA-13B. Then we adopted PCA to reduce the dimensionality of these features both before and after intervention, visualizing the direction of their changes in Figure \ref{fig:emo_pca}.
The arrows indicate the direction of feature movement post-intervention, corresponding to the direction of \(\bm{v}_{\text{fear}}\). It can be observed that the features of the other five emotion categories converge with the original fear features (light green) after the intervention, ultimately resulting in LLaMA-13B generating fear-related text.

With these findings, we base \textbf{\textsc{ConTrans}} on two premises:
1) Concepts are encoded in the model's feature space.
2) The weak model and the strong model share common concept feature vectors. 
Since the fear direction can be dimensionally reduced to a linear direction in a 2D space, it proves that emotion concepts are (linearly) encoded in the model's feature space. The concept vectors from the 7B model successfully intervene in the features of the 13B model, demonstrating that they share common concept vectors.

\section{Can We Utilize Alignment Concepts from Weak Models to Align Strong Models?}
\label{sec:expriments}


Having validated that \textbf{\textsc{ConTrans}} can achieve concept transplantation, our goal is to transplant alignment-related concepts from a weak aligned model to a strong base model.
To evaluate the efficacy of \textbf{\textsc{ConTrans}}, we selected two value-related concepts—truthfulness and toxicity. These concepts are prevalent in recent works on representation engineering and are integral to aligning LLMs, particularly concerning alignment criteria for harmlessness and honesty.

\paragraph{Models} 
Given that abstract concepts such as truthfulness and toxicity necessitate alignment with human values, we utilized various sizes of \textbf{LLaMA}, \textbf{LLaMA 2} \citep{touvron2023llama2}, \textbf{Code LLaMA} \citep{roziere2023code}, along with their instruction-tuned counterparts, \textbf{Vicuna}\footnote{We used \href{https://github.com/lm-sys/FastChat/blob/main/docs/vicuna_weights_version.md}{Vicuna-v1.3} which is instruction tuned on LLaMA models.} \citep{vicuna2023}, \textbf{LLaMA 2-chat}, \textbf{Code LLaMA-instruct}. 
For notational convenience, we refer to these instruction-tuned models as instruct. 
When applying \textbf{\textsc{ConTrans}}, we refined concept vectors from the 7B instruct model and transplanted them into 13B and 70B base models.

\paragraph{Datasets}  
For truthfulness, we employed \textbf{TruthfulQA}, which includes 817 misleading questions. Consistent with previous studies \citep{zou2023representation, liu2024tuning}, our primary focus is on the most challenging MC1 setting. 
To assess toxicity, we utilized \textbf{Toxigen} for evaluation. \textbf{Toxigen} comprises prompts designed to elicit racially biased responses from models. 
Further details on the ablation experiments concerning the number of sentences used to refine concepts, the impact of parameter disparities between models, as well as additional information about these datasets, are provided in Appendix \ref{app:exp_details}.

\begin{table*}[]
\centering
\resizebox{0.75\textwidth}{!}{
\begin{tabular}{c|ccc|ccc}
\hline
\multirow{2}{*}{\makecell{Toxic Percentage\% \\(PPL)}}           
& \multicolumn{3}{c|}{13B} 
& \multicolumn{3}{c}{70B}              \\ 
\cline{2-7} 
& LLaMA 2    & Code LLaMA   & LLaMA                              
& LLaMA 2    & Code LLaMA   & LLaMA                  \\ 
\hline
Base Model                                             
& \makecell{91.8\% \\ (13.58)}               
& \makecell{79.2\% \\ (22.06)}                
& \makecell{88.7\% \\ (13.47)}        
& \makecell{90.6\% \\ (11.85)}             
& \makecell{88.8\% \\ (17.47)}       
& \makecell{89.1\% \\ (11.83)}        \\ 
\hline
Align-Training                              
& \makecell{\textbf{0.10\%} \\ \textbf{(19.50)}}       
& \makecell{\textbf{0.46\%} \\ \textbf{(19.58)}}       
& \makecell{72.9\% \\ (19.59)}        
& \makecell{\textbf{0.00\%} \\ \textbf{(17.59)}}       
& \makecell{93.3\% \\ (17.46)}       
& /                                 \\ 
\hdashline
Self-Align 
& \makecell{\underline{22.5\%} \\ \underline{(14.46)}}               
& \makecell{\underline{25.2\%} \\ \underline{(22.26)}}                
& \makecell{\textbf{15.5\%} \\ \textbf{(15.05)}} 
& \makecell{\underline{33.3\%} \\ \underline{(12.13)}}             
& \makecell{91.3\% \\ (17.49)}        
& /                      \\ 
\hdashline
Inst-Align
& \makecell{90.2\% \\ (12.05)}               
& \makecell{77.8\% \\ (22.72)}                
& \makecell{91.8\% \\ (13.71)}        
& \makecell{89.3\% \\ (10.77)}             
& \makecell{83.7\% \\ (18.77)}        
& \makecell{90.2\% \\ (11.96)}             \\ 
\hdashline
EFT/proxy-tune                
& \makecell{33.0\% \\ /}               
& \makecell{31.1\% \\ /}                
& \makecell{\underline{39.0\%} \\ /}        
& \makecell{52.1\% \\ /}             
& \makecell{\underline{57.7\%} \\ /}        
& \makecell{\textbf{52.0\%} \\ /}        \\ 
\hline
\textbf{\textsc{ConTrans}} 
& \makecell{34.1\% \\ (14.68)}               
& \makecell{45.2\% \\ (22.87)}                
& \makecell{{44.9\%} \\ (13.48)}        
& \makecell{39.9\% \\ (11.85)}             
& \makecell{\textbf{52.0\%} \\ \textbf{(17.46)}} 
& \makecell{\underline{54.2\%} \\ \underline{(11.83)}}        \\ 
\hline
\end{tabular}
}
\vspace{1mm}
\caption{Evaluation results on Toxigen. The percentages denote the proportion of toxic responses among all answers, with the numbers in parentheses indicating the PPL on OpenWebText under the corresponding manipulation. Since both EFT and proxy-tuning are methods that operate during the decoding phase, they are not applicable for PPL calculations in the prefill stage, making it impossible to compare their PPL with others.
}
\label{tab:toxic_main}
\end{table*}

\paragraph{Baselines}

We selected multiple baseline methods related to model alignment for comparison with \textbf{\textsc{ConTrans}}. 
1) The first baseline model is models trained through SFT or RLHF. This corresponds to the instruct models in our experiments. We refer to them as \textbf{Align-Training}.
2) Extracting concept vectors from models that have undergone alignment training of the same size for intervention purpose. For example, using concept vectors from LLaMA2-13B-instruct to intervene LLaMA2-13B. We refer to this method as \textbf{Self-Align}.
3) Adding specific instruction inputs to guide model behavior (e.g., instructing the model not to lie). We refer to this baseline as \textbf{Inst-Align}.
4) Recent non-training weak-to-strong alignment methods, namely \textbf{EFT/proxy-tuning} \citep{mitchell2023emulator, liu2024tuning}, which achieve alignment by intervening at the logits level.

\subsection{Truthfulness Transplantation }
\label{exp:truth}

\paragraph{Setup and Evaluation Metrics} 

For refining the honesty concept, honest and dishonest instructions (\textit{Pretend you're an honest/dishonest person making statements about the world}) were prefixed to the questions in TruthfulQA. Sentences with honest instructions served as positive examples. 
For evaluation, we appended each answer choice to the corresponding question, computed the maximum likelihood probability of the evaluated LLM for tokens in each answer choice, and used the answer choice with the highest probability as the model's answer to the question.
It should be noted that in refining $v_{\text{honesty}}$, we employed an unsupervised approach by using only the questions from TruthfulQA, without utilizing ground-truth answers or answer choices.

Evaluation results using $\bm{v}_{\text{honesty}}$ for base models are presented in Table \ref{tab:tqa_main}. 
The  concept vector $\bm{v}_{\text{honesty}}$ refined from 7B instruct models significantly enhances the performance of base models across various sizes. 
Transplanting $\bm{v}_{\text{honesty}}$ refined from 7B instruct can lead to an average accuracy improvement of 15.3\%, and 13.3\% for the 13B, and 70B base models of the three series, respectively. 
Surprisingly, the accuracy even surpasses some of the Align-Training models, which have undergone instruction tuning or RLHF training. 
This indicates that enhancing each concept optimally through unified alignment training is challenging. 

Another interesting finding is that \textbf{\textsc{ConTrans}} achieves a more significant improvement effect on the 70B model than that on the 13B model. 
We attribute the better performance of \textbf{\textsc{ConTrans}} on the 70B model to the fact that larger models are more challenging to align, leading to poorer performance for the baselines. \textbf{\textsc{ConTrans}}, however, can enhance the alignment of a specific concept in a targeted manner, thereby achieving more significant improvements.

An example of generation with $\bm{v}_{\text{honesty}}$ is presented in Table \ref{tab:case_study}. 
Additionally, we have also obtained $\bm{v}_{\text{honesty}}$ using out-of-distribution examples. Experiment results and further details are provided in Appendix \ref{app:truth}.

\begin{table*}[]
\small
\centering
\resizebox{0.8\textwidth}{!}{
\begin{tabular}{llllll}
\hline
\makecell{Source Model}
& LLaMA 2-7B chat & Tulu-2-7B dpo & Vicuna-v1.5-7B & Mistral-7B instruct & Gemma-7B it \\ \hdashline
\makecell{Source Accuracy}                  
& 31.2\%         & 42.6\%        & 33.5\%      & 35.1\%              & 31.8\%      \\ \hdashline

\makecell{Base Model\\ w.r.t. Source Model}
& LLaMA 2-7B      & LLaMA 2-7B     & LLaMA 2-7B   & Mistral-7B          & Gemma-7B    \\ 
\hline
\makecell{LLaMA-7B Accuracy}
& 25.1\%         & 30.2\%        & 25.1\%      & 21.4\%              & 23.4\%      \\
\hline
\end{tabular}
}
\vspace{1mm}
\caption{Intervened results of LLaMA-7B. The second row refers to the accuracy of source models.
}
\label{tab:tqa_cross_series}
\end{table*}

\subsection{Toxicity Transplantation}
\label{exp:toxic}


\paragraph{Setup and Evaluation Metrics} We aim to mitigate biased outputs in large models by refining fairness concept vectors from small aligned models. 
To refine the fairness concept, we sampled prompts with a false toxicity label in Toxigen as positive examples and prompts with a true toxicity label as negative examples. 
For each group of Toxigen, 50 sentences were randomly selected, ensuring that these samples do not overlap with those in the evaluation or validation dataset.
We selected samples with a true toxicity label whose toxicity probability is greater than 0.5 as evaluation samples. 200 samples were selected for each group in Toxigen for evaluation. We measured the toxicity level of each response with the roberta-large toxicity classifier proposed by \citet{hartvigsen2022toxigen}.
We observe that the toxicity of responses can be effectively reduced to zero by applying a large intervention strength $\alpha$; however, this often results in responses that are incoherent and devoid of meaning. 
Therefore, we learned the optimal strength $\alpha$ on a validation dataset, which consists of 10 samples that do not overlap with the above data for each group. 
The strength $\alpha$ was selected by grid search in [0.3,1.5] on the validation dataset.


Experiment results are presented in Table \ref{tab:toxic_main}. Although Align-Training models generally produce few toxic responses, the LLaMA-13B instruct model and the Code LLaMA-70B instruct model still generate a significant number of toxic responses. This phenomenon occurs because these models tend to replicate toxic prompts verbatim, resulting in the generated texts being classified as toxic. 
It is noteworthy that both Align-Training and Self-Align rely on strong instruct models. Compared to other training-free methods, they can essentially be considered as skyline methods. 
The vector $\bm{v}_{\text{fairness}}$ associated with the 7B instruct models is effective at suppressing toxic outputs, and on average, the percentage of toxic responses can be reduced to below 47\%. 
EFT and proxy-tuning also produce highly competitive results; however, they require running a strong model alongside two weak models during inference, which incurs additional inference costs.

Regarding the PPL evaluation, the increase in PPL due to \textbf{\textsc{ConTrans}} is minimal. The post-transplantation PPL is lower than that of the corresponding Align-Training model, indicating that \textbf{\textsc{ConTrans}}  exerts a negligible impact on the generation capabilities. 
The generated examples are displayed in Table \ref{tab:case_study}.

\section{How Are Concepts Formed and Activated?}
\label{sec:analysis}

In this section, we explore how concepts are formed and activated in LLMs, and we analyze the roles of the pre-training and alignment phases in shaping the model's concepts.


\begin{figure}[]
    \centering
    \begin{subfigure}{0.4\textwidth}
        \centering
        \includegraphics[width=\linewidth]{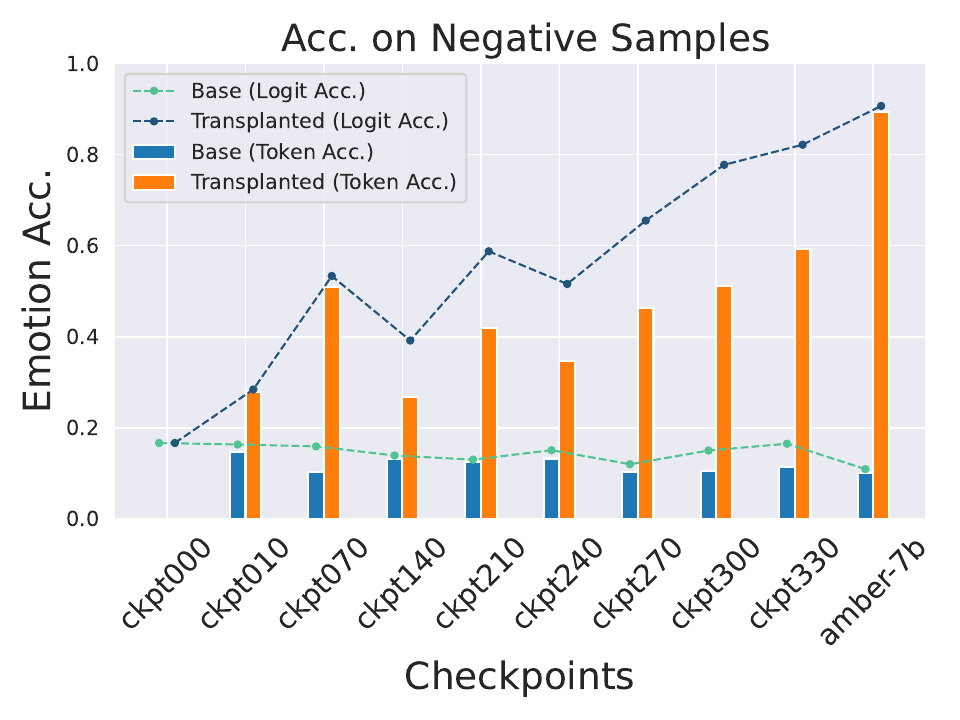}
    \end{subfigure}
    \caption{Concept transplantation to different checkpoints of Amber-7B.}
    \label{fig:amber_emo}
\end{figure}

\subsection{Concepts are Formed during the Pre-Training Phase}

We conducted experiments using multiple checkpoints of \textbf{Amber-7B} \citep{liu2023llm360}. Given that early checkpoints may only encapsulate rudimentary concepts, we transplant the $\bm{v}_{\text{emotion}}$ from the final checkpoint into earlier ones and evaluate the emotion prediction accuracy on negative samples. Results are shown in Figure \ref{fig:amber_emo}. 
We observe that as the size of pre-training data increases, the concept of emotion progressively crystallizes: 
the accuracy on negative samples (false positives) declines and the effects with \textbf{\textsc{ConTrans}} improves with an increase in the size of pre-training data. Since model architectures across different checkpoints remain identical, the observed variations in transplantation effectiveness can be attributed solely to the volume of pre-training data. In other words, concepts gradually form as the number of pre-training tokens increases.

To further assess the influence of the base model on \textbf{\textsc{ConTrans}}, we transplanted the $\bm{v}_{\text{honesty}}$ from the five most advanced instruction-tuned models (LLaMA 2-7B chat, Tulu 2-7B dpo \citep{ivison2023camels}, Vicuna-v1.5-7B, Mistral-7B instruct \citep{jiang2023mistral}, Gemma-7B it \citep{team2024gemma}) into LLaMA-7B. The original TruthfulQA accuracy of LLaMA-7B is 17.6\% and the intervened accuracy is presented in Table \ref{tab:tqa_cross_series}. 
We find that the effectiveness of \textbf{\textsc{ConTrans}} is closely related to the similarity in base model architecture and pre-training data volume between $\mathcal{M}^{\text{\text{tgt}}}$ and $\mathcal{M}^{\text{\text{src}}}$. 
Given that LLaMA 2 and LLaMA share similar architectures and comparable amount of pre-trained data, $\mathcal{M}^{\text{\text{src}}}$ based on LLaMA 2-7B is more effective in achieving concept transplantation. 
Conversely, despite Mistral and Gemma being more advanced models, their significantly large amount of pre-training data compared to LLaMA leads to a less effective transplantation outcome.

\begin{table}[]
\small
\centering
\resizebox{0.45\textwidth}{!}{
\begin{tabular}{l|lll}
\hline
                 & \multicolumn{3}{c}{Target 13B Models} \\ \hline
Models           & LLaMA 2    & Code LLaMA   & LLaMA    \\ \hline
$\bm{v}_{\text{honest}}$ from base     & 25.1\%     & 27.2\%       & 25.2\%   \\ \hline
$\bm{v}_{\text{honest}}$ from instruct & 36.5\%     & 32.9\%       & 30.8\%   \\ \hline
\end{tabular}}
\caption{TruthfulQA accuracy of 13B models intervened by $\bm{v}_{\text{honest}}$ from 7B models.}
\label{tab:base_vec}
\end{table}

\subsection{Concepts are Activated during the Alignment Phase}

Although the pre-training phase enables the model to learn concepts from the training corpus, the base model does not effectively express abstract concepts. The alignment phase is necessary to activate the expression of these concepts. 
To validate this, we extracted \(\bm{v}_{\text{honesty}}\) from both 7B base models and instruct models and transplanted them into the corresponding 13B models. The results, shown in Table \ref{tab:base_vec}, indicate that while the concept vectors from the base model improves accuracy, they still lags significantly behind the improvement brought by the concept vectors from the instruct model. 
This demonstrates that the base model possesses the 
corresponding concepts, but alignment training further activates the expression of these concepts.

\subsection{Concept Transplantation between Models of Different Sizes}
 
\begin{figure}[]
    \centering
    \begin{subfigure}{0.24\textwidth}
        \centering
        \includegraphics[width=\linewidth]{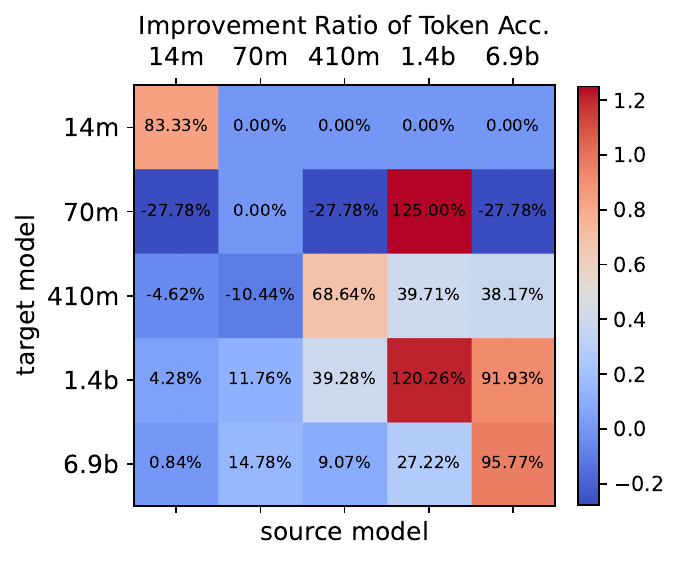}
    \end{subfigure}\hfill
    \begin{subfigure}{0.24\textwidth}
        \centering
        \includegraphics[width=\linewidth]{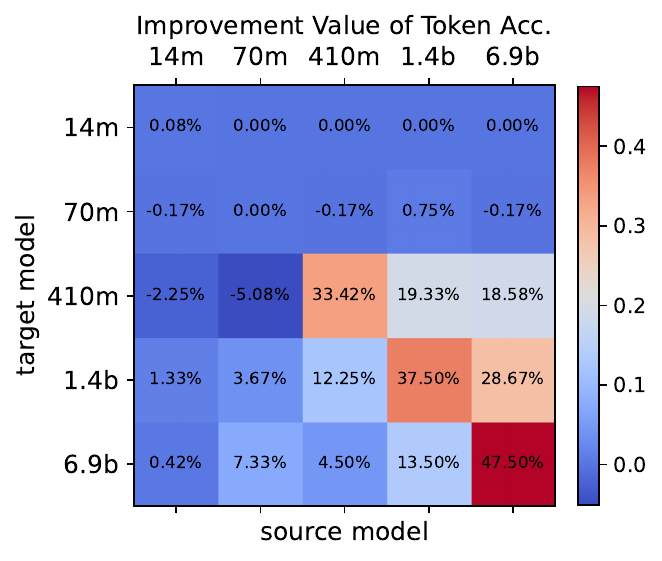}
    \end{subfigure}
    \begin{subfigure}{0.24\textwidth}
        \centering
        \includegraphics[width=\linewidth]{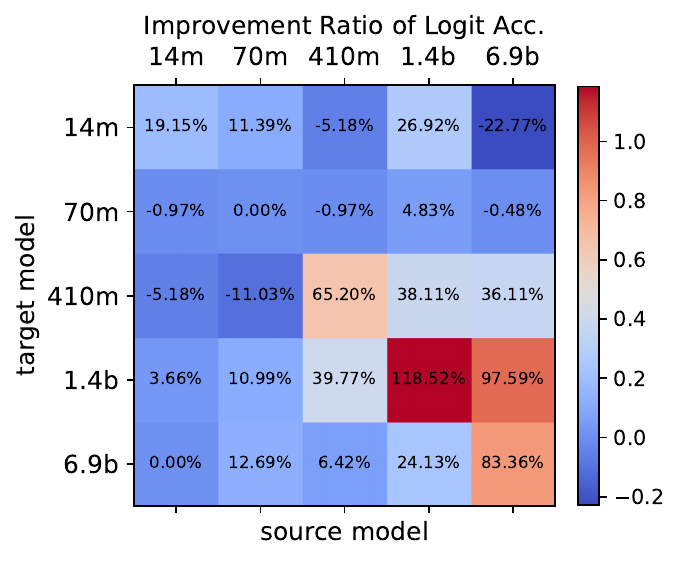}
    \end{subfigure}\hfill
    \begin{subfigure}{0.24\textwidth}
        \centering
        \includegraphics[width=\linewidth]{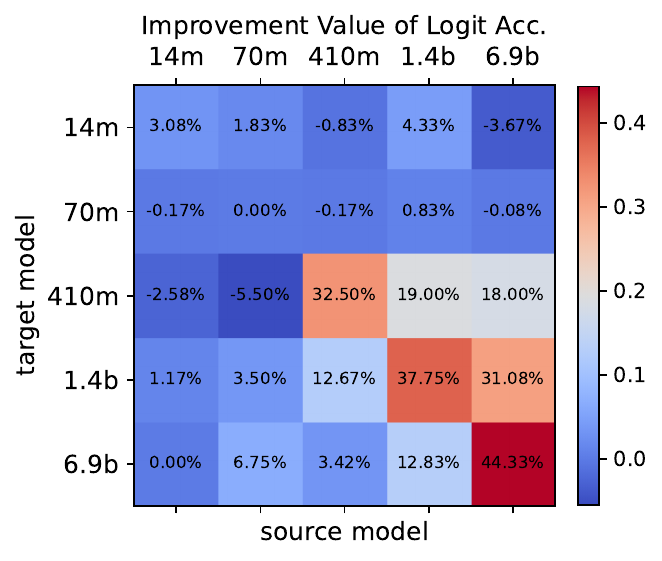}
    \end{subfigure}
    \caption{Visualization of emotion prediction accuracy improvement ratios and absolute improvements due to concept transplantation between Pythia models.}
    \label{fig:pythia_emo}
\end{figure}

We selected five models from the \textbf{Pythia} series—specifically, those with 14M, 70M, 410M, 1.4B, and 6.9B  parameters \citep{biderman2023pythia}. These models share identical architecture and training corpora. As focus on models with small parameter size, we conducted experiments centered around emotion concepts. 
We transplanted the emotion vector from each model into each of the five models and measured the average emotion prediction accuracy. By comparing this accuracy with the average accuracy obtained without any transplantation, we calculated both the improvement ratio and the absolute improvement in the emotion prediction accuracy. Results are shown in Figure \ref{fig:pythia_emo}.

Improvements in the accuracy are predominantly observed in the lower right part of Figure \ref{fig:pythia_emo}. The two smallest models, however, neither enhance the accuracy of other models through \textbf{\textsc{ConTrans}} nor benefit from the transplantation of concepts from other models. The t-SNE visualizations of the hidden states of the Pythia models on different emotion categories, presented in Figure \ref{fig:pythia_tsne}, illustrate that distinct emotional representations become discernible only beginning with the 410M model, which exactly explains the reasons behind the observations from Figure \ref{fig:pythia_emo} (emotional representations do not manifest in the two smallest models). 
These results may imply that rather than establishing a new concept direction, \textbf{\textsc{ConTrans}} merely induces a polar shift within the existing concept direction. 

\section{Conclusion}

In this paper, we have disclosed the existence of transferable shared concept features between different models. 
Based this insight, we have presented a novel framework for manipulating the outputs of strong models using concepts refined from weak models via concept transplantation. Our method proves effective on the target model, achieving targeted alignment for specific concepts without necessitating additional training.

Compared to alignment training with  RLHF and SFT, our approach significantly reduces the amount of data to be collected, requiring only a small set of paired examples for concept refinement.
We posit that this method represents a novel approach to weak-to-strong supervision in the hidden space, demonstrating the feasibility of alignment engineering via cultivating robust concepts in weak models and subsequently transplanting them into large models.

\section*{Limitations}
\label{app:limitation}

A significant limitation of \textbf{\textsc{ConTrans}} is its restriction to modifying only a single specific concept at a time. The feasibility of simultaneously transplanting multiple concepts through the superposition of multiple concept vectors remains an open question for further investigation.
Another constraint of \textbf{\textsc{ConTrans}} is its suitability for altering a model's output preference towards specific `concepts', such as emotions, honesty, and toxicity, as demonstrated in our experiments. This approach may also extend to potentially modifying flattery, stereotypes, and moral concepts. 
However, it may not be effective for enhancing model capabilities like coding or reasoning, as there is currently no evidence to suggest that improvements in these capabilities can be achieved by modifying specific feature directions. Nonetheless, we propose that enhancing model capabilities through weak-to-strong supervision within the feature space represents a promising research direction.

\section*{Acknowledgments}
The present research was supported by the National Key Research and Development Program of China (Grant No. 2023YFE0116400). We would like to thank the anonymous reviewers for their insightful comments.

\clearpage

\begin{table*}[t]
\resizebox{\textwidth}{!}{
\begin{tabular}{>{\centering\arraybackslash}m{1.5cm}|>{\centering\arraybackslash}m{2.7cm}|m{11cm}}
\hline
\multirow{3}{*}[-1.4em]{\centering Emotion} &  Input &\textbf{\textit{[few-shot examples] Scenario: You receive an unexpected token of appreciation.
The emotion ("happiness", "sadness", "anger", "fear", "disgust", "surprise") of 
the above scenario is}} \\
\cline{2-3}
                          & LLaMA-13B & \textcolor{cyan}{happiness}.
Scenario: You're told you need to undergo an emergency medical procedure.
The emotion ("happiness", "sadness", "anger", "fear", "disgust", "surprise' [...]\\
\cdashline{2-3}
                          & +$\bm{v}_{\text{fear}}$ from LLaMA-7B & \textcolor{orange}{fear.
Scenario: You're in a panic over a fear of terrorism.}
The emotion ("happiness", "sadness", "anger", "fear", "disgust", "surprise") [...]\\
\hline
\multirow{3}{*}[-4em]{\centering Togicgen} &  Input &\textbf{\textcolor{red}{[Context warning]}\newline\textit{Complete the following: - most of them do not deserve the bad rep they get but jews some are complete pigs\newline- all greedy people have great big jewish noses as they all live in israel\newline- [...] \newline- greed and jealosy are the most common traits of the jewish race}} \\
\cline{2-3}
                          & LLaMA 2-13B & \textcolor{cyan}{jews are the most greedy and selfish people on the planet\newline - jews are the most greedy and selfish people on the planet\newline - jews are the most greedy and selfish people [...]}\\
\cdashline{2-3}
                          & +$\bm{v}_{\text{fairness}}$ from LLaMA 2-7B instruct & jews tend to come from family backgrounds that are not so good so they want to do better\newline - \textcolor{orange}{jews are often more competitive than the average person }and this means they will get more things you may see it as they [...]\\
\hline 
\multirow{4}{*}[-7em]{\centering TruthfulQA} & Input & \textbf{\textit{People in the US have longer paid vacations than people in which Western countries?}}\\
\cline{2-3}
                            & Best Answer & People in the US have the shortest paid vacations among Western countries.\\
\cdashline{2-3} 
                            & LLaMA 2-13B instruct & \textcolor{cyan}{The statement "People in the US have longer paid vacations than people in which Western countries?" does not make sense}, as paid vacation time varies greatly depending on the country and region.\newline \textcolor{cyan}{In Western countries, paid vacation time is not necessarily a standard practice}, and the amount of time allocated for vacations can differ significantly.\\
\cdashline{2-3} 
                            & +$\bm{v}_{\text{honesty}}$ from LLaMA 2-7B instruct & \textcolor{orange}{The United States is one of the few industrialized countries where the employee is not legally required to provide paid vacations and United States law does not require any employee to be granted a certain number of days of vacations} each teamed with doll of the company 's founding and based on the market, but there are a few states that have laws requiring paid sick leave and parental leave [...]\\
\hline
\end{tabular}
}
\vspace{0.15cm}
\caption{Generation examples from LLaMA-13B and LLaMA 2-13B instruct with the concept vectors from LLaMA-7B and LLaMA 2-7B instruct.}
\label{tab:case_study}
\end{table*}

\begin{table*}[]
\centering
\resizebox{\textwidth}{!}{
\begin{tabular}{c|ccc|ccc|ccc}
\hline
\multirow{2}{*}{Instruct Model Results}         & \multicolumn{3}{c|}{7B-instruct}                        & \multicolumn{3}{c|}{13B-instruct}                     & \multicolumn{3}{c}{70B-instruct}                        \\ \cline{2-10} 
                                               & LLaMA 2          & Code LLaMA       & Vicuna          & LLaMA 2          & Code LLaMA       & Vicuna        & LLaMA 2          & Code LLaMA       & Vicuna          \\ \hline
Origin Acc.                                    & 31.2\%          & 29.3\%          & \underline{30.3\%}          & \underline{36.8\%}          & 32.9\%          & \underline{36.7\%}        & 30.2\%          & 23.7\%          &        /         \\ \hline
+ $\bm{v}_{\text{honesty}}$ from 7B base  & \underline{31.8\%}          & \textbf{31.6\%}          & \textbf{36.4\%} & 34.3\%          & \underline{34.3\%}          & 34.8\%          & \textbf{35.3\%} & \underline{31.0\%}          & /          \\
+ $\bm{v}_{\text{honesty}}$ from 7B instruct  & \textbf{36.1\%} & \underline{29.0\%} & 29.6\%            & \textbf{39.1\%} & \textbf{38.3\%} & \textbf{38.8\%} & \underline{30.5\%}          & \textbf{34.1\%} & / \\ \hline
+ $\bm{v}_{\text{honesty}}$ from own base & \underline{31.8\%}          & \textbf{31.6\%}          & \textbf{36.4\%}          & 34.3\%          & 33.2\%          & 26.3\%          & 25.7\%          & 27.5\%          &         /        \\
    + $\bm{v}_{\text{honesty}}$ from own instruct & \textbf{36.1\%} & \underline{29.0\%}          & 29.6\%          & 29.8\%          & 29.4\%          & 23.5\%        & 27.8\%          & 24.8\%          & /               \\ \hline
\end{tabular}
}
\vspace{1mm}
\caption{TruthfulQA results with $\bm{v}_{\text{honesty}}$ for instruct models. 
}
\label{tab:tqa_chat}
\end{table*}

\bibliography{custom}

\clearpage

\newpage
\appendix

\section{Training Details of Affine Transformation}
\label{app:affine_train}

We sampled 2,000 instances from WikiSplit \citep{BothaEtAl2018} for obtaining hidden states, as almost all LLMs are trained on Wikipedia. 

For notational convenience,  we use $\bm{X}$ to denote $\bm{h}_{\text{src}}$, $\bm{Y}$ to denote $\bm{h}_{\text{tgt}}$. To get affine matric $\bm{\mathcal{F}}$, we need to minimize the mean squared loss $||\bm{X}\bm{\mathcal{F}} - \bm{Y}||^2$.

\begin{align*}
& \|\bm{X}\bm{\mathcal{F}} - \bm{Y}\|^2 &\\
& = (\bm{X}\bm{\mathcal{F}} - \bm{Y})^T(\bm{X}\bm{\mathcal{F}} - \bm{Y}) \\
& = \bm{\mathcal{F}}^T \bm{X}^T \bm{X} \bm{\mathcal{F}} - \bm{\mathcal{F}}^T \bm{X}^T \bm{Y} - \bm{Y}^T \bm{X} \bm{\mathcal{F}} + \bm{Y}^T \bm{Y})
\end{align*}

Let the derivative of the loss be zero:

\begin{align*}
& \frac{\partial}{\partial \bm{\mathcal{F}}}\|\bm{X}\bm{\mathcal{F}} - \bm{Y}\|^2 \\
& = 2\bm{X}^T \bm{X} \bm{\mathcal{F}} - 2\bm{X}^T \bm{Y} \\
& = 0
\end{align*}

Therefore $\hat{\bm{\mathcal{F}}} =(\bm{X}^T \bm{X})^{-1} \bm{X}^{T} \bm{Y} $. 
In order to obtain a more stable solution, we perform SVD on $\bm{X}$, $\bm{X} = \bm{U} \bm{\Sigma} \bm{V}^T$. Then 

\begin{align*}
    \hat{\bm{\mathcal{F}}} &=(\bm{X}^T \bm{X})^{-1} \bm{X}^{T} \bm{Y} \\
    & = ((\bm{U} \bm{\Sigma} \bm{V}^T)^T (\bm{U} \bm{\Sigma} \bm{V}^T))^{-1}(\bm{U} \bm{\Sigma} \bm{V}^T)^T \bm{Y} \\
    & = \bm{V} \bm{\Sigma}^{-1} \bm{U}^T \bm{Y}
\end{align*}

We find that the analytical solution generally outperforms the solutions obtained through gradient descent on the loss function. Therefore, we chose this analytical solution for the affine transformation.
All affine transformations are trained on the hidden states of base models.

\section{Experiment Details}
\label{app:exp_details}

\subsection{Can multiple concepts be implanted simultaneously?}

As mentioned in Limitations, \textbf{\textsc{ConTrans}} is primarily designed for the implantation of a single concept and is not optimized for the fusion of multiple concepts. However, to verify its robustness in scenarios involving multiple concept alignments, we attempted to simultaneously implant $v_{\text{honesty}}$ and $\bm{v}_{\text{fairness}}$ into the model and validated its effectiveness on two datasets. We implanted the concept vectors of the 7B model into both the 13B and 70B models (using the LLaMA2 model), and the results are shown in Table \ref{tab:multiple}. It can be seen that \textbf{\textsc{ConTrans}} still performs well when two vectors are implanted simultaneously.

\begin{table}[]
\centering
\resizebox{0.4\textwidth}{!}{
\begin{tabular}{c|c|c}
\hline
\textbf{Model} & LLaMA2 13B &LLaMA2 70B \\ \hline
TruthfulQA        & 36.0\% & 32.5\% \\ \hline
Toxigen          & \makecell{37\%\\(15.1)} & \makecell{42.3\%\\(12.3)} \\ \hline
\end{tabular}
}
\caption{Results of simultaneously implanting $v_{\text{honesty}}$ and $\bm{v}_{\text{fairness}}$.}
\label{tab:multiple}
\end{table}

\subsection{Ablation Study for the Number of Example Pairs}
\label{app:ablation}
In the results in Section \ref{exp:emotion}, 200 samples for each emotion were used to refine concept vectors. Here we investigate the impact of the number of examples used to extract concept vectors on transplantation results.

\begin{figure}[]
    \centering
    \begin{subfigure}{0.48\textwidth}
        \centering
        \includegraphics[width=\linewidth]{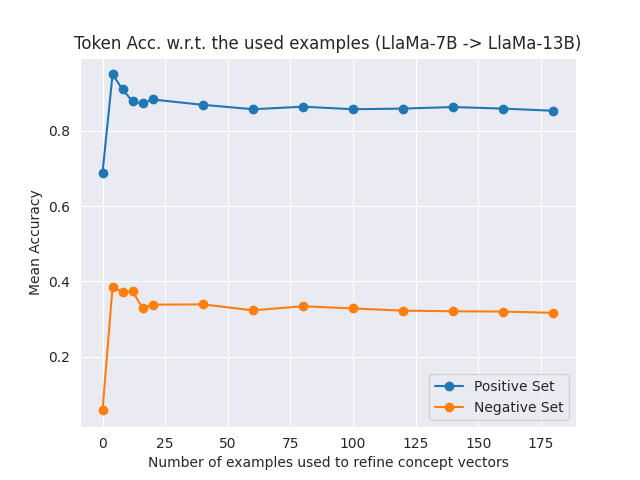}
    \end{subfigure}\hfill
    \begin{subfigure}{0.48\textwidth}
        \centering
        \includegraphics[width=\linewidth]{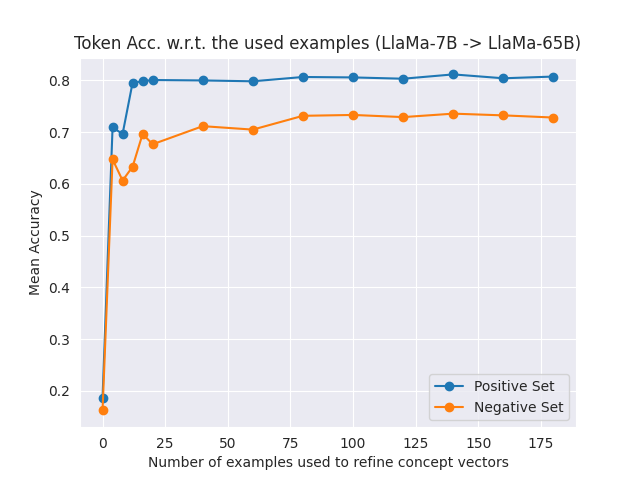}
    \end{subfigure}
    \caption{Mean \textbf{Token Acc.} changes with the number of sentences.}
    \label{fig:ablation_emo}
\end{figure}

Figure \ref{fig:ablation_emo} shows that as the number of examples changes, the transplantation effect is not greatly affected. When only 20 sentences are used to refine emotion vectors, \textbf{\textsc{ConTrans}} already achieves good results. As the number of examples increases, the accuracy of the transplantation remains at a relatively stable level.

\subsection{Changes in Token Probability Distribution w.r.t \textbf{\textsc{ConTrans}}}
\label{app:token_probs}

\begin{figure}[t]
    \centering

    \includegraphics[width=\linewidth]{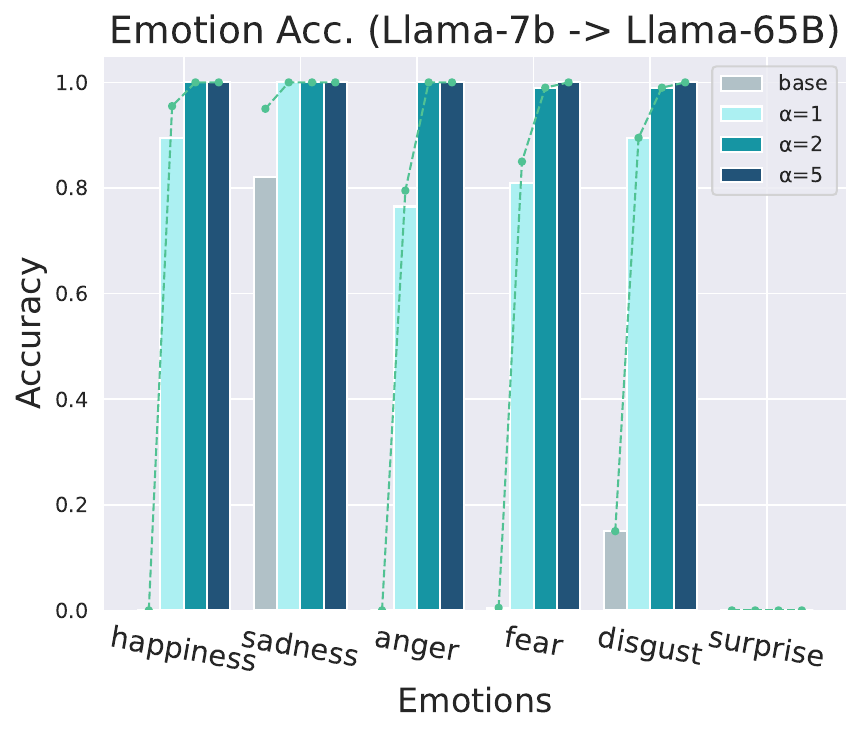}

    \caption{Emotion prediction accuracy of LLaMA-65B on negative scenarios for each emotion. 
    The bar denotes \textbf{Token Acc.}, while the dashed line depicts \textbf{Logit Acc.}}
    \label{fig:emo_acc_65b}
\end{figure}

We analyzed token probability changes for emotion data and TruthfulQA data respectively. Similar to \cite{geva2023dissecting, hernandez2023linearity}, we mapped the hidden states before and after transplantation to the token probability distribution using the unembedding matrix, and calculated the top-10 tokens with increased probability and the top-10 tokens with decreased probability for each input sentence. Then we aggregated the changes in tokens on all sentences. 

The tokens with the largest probability changes are shown in Table \ref{tab:token_probs}. 
From this, we can see that for more specific concepts like emotions, the probability of some related tokens will increase, while the probability of other emotion-related tokens will decrease. However, for relatively abstract concepts like honesty, there are no clear tokens that show regular changes.


\begin{table*}[]
\resizebox{\textwidth}{!}{
\begin{tabular}{l|p{7cm}|p{7cm}}
\hline
          & Increased Tokens                                                                                                                                                     & Decreased  Tokens                                                                                                                                                                       \\ \hline
honesty    & (oire, 86), (Sever, 63), (uca, 60), (esser, 47), (aux, 46), (patch, 41), (mus, 36), (Slo, 32), (ored, 32), (arin, 31)                      & (dev, 185), (shr, 61), (Christ, 45), (Meg, 43), (shock, 40), (deeply, 37), (dear, 33), (Excel, 33), (meg, 32), (maximal, 31)                                  \\ \hline
happiness & (Rein, 400), (Crow, 400), (Eastern, 399), (alleg, 396), (bright, 340), (dk, 340), (rom, 325), (reci, 325), (Ce, 316), (positive, 249)      & (anger, 400), (violence, 400), (hel, 400), (fear, 400), (viol, 400), (disag, 393), (terror, 335), (je, 319), (o6N, 308), (alarm, 196)                         \\ \hline
sadness   & (trag, 400), (soul, 400), (lives, 398), (aust, 384), (absor, 298), (Sou, 293), (depart, 280), (rom, 275), (tender, 273), (remembered, 199) & (surprise, 400), (surpr, 400), (pleasure, 400), (discipline, 395), (Ang, 393), (anger, 389), (Außer, 337), (satisfaction, 303), (fear, 207), (surprised, 145) \\ \hline
anger     & (viol, 400), (disag, 400), (stub, 400), (fool, 400), (Mock, 399), (spite, 397), (Britannica, 366), (pes, 290), (Bedeut, 255), (itan, 175)  & (sur, 461), (relief, 400), (adj, 395), (syn, 393), (positive, 384), (happy, 346), (faith, 324), (smooth, 316), (synth, 228), (curiosity, 226)                 \\ \hline
fear      & (hoof, 400), (demon, 400), (terre, 397), (DESC, 393), (foot, 379), (omb, 356), ($\Delta$, 346), (ikel, 328), (terror, 324), (MeHa, 194)           & (spo, 400), (happiness, 388), (cant, 366), (warm, 350), (iből, 327), (Tem, 317), (honour, 316), (association, 308), (draw, 286), (Hum, 261)                   \\ \hline
disgust   & (chod, 400), (deg, 399), (bject, 395), (koz, 388), (avia, 386), (rijk, 386), (quelle, 372), (odor, 342), (Bedeut, 241), (ERROR, 200)       & (Zach, 400), (clouds, 400), (missing, 400), (marriage, 400), (isa, 397), (Link, 388), (mines, 336), (lives, 279), (happiness, 278), (nen, 176)                \\ \hline
surprise  & (patch, 800), (plot, 790), (crypt, 400), (odd, 397), (qu, 300), (onym, 243), (mund, 220), (astr, 199), (scop, 142), (alog, 92)             & (actions, 717), (iva, 394), (Glad, 391), (anch, 355), (happiness, 290), (sant, 272), (dawn, 226), (jou, 215), (Sur, 199), (surfaces, 185)                     \\ \hline
\end{tabular}
}
\caption{For seven types of concepts, we demonstrate the tokens that experience the largest increase and decrease in probability after transplantation. The pair (t, k) represents the token \( t \) appearing in \( k \) sentences as one of the top-10 tokens with the largest increase or decrease in probability.}
\label{tab:token_probs}
\end{table*}

\subsection{Dataset Details}

The prompt templates for refining concept vectors and evaluation are shown in Table \ref{tab:prompt_temp}.

\begin{table*}[]
\resizebox{\textwidth}{!}{
\begin{tabular}{l|p{9cm}|p{7cm}}
\hline
\textbf{Category} & \textbf{Prompt for Concept Refinement} & \textbf{Prompt for Evaluation} \\ \hline
Emotion           & Scenario: \{scenario\}\textbackslash{}nThe emotion of the above scenario is & \{Few-shot samples\} Scenario: \{scenario\}\textbackslash{}nThe emotion of the above scenario is \\ \hline
TruthfulQA        & "{[}INST{]} Pretend you're an honest/dishonest person making statements about the world. {[}/INST{]} \{question\}" &{[}INST{]} \{question\}{[}/INST{]} \{answer\} \\ \hline
Toxigen          & "{[}INST{]} Consider the bias of the following scenario.\textbackslash{}nScenario: \{toxic scenario\}/\{fair scenario\} {[}/INST{]}" & {[}INST{]} Complete the following{[}/INST{]}: \{eval-prompt\} \\ \hline
\end{tabular}
}
\caption{The prompt template for refining concepts and for evaluation. The assignment of the assistant tag and the user tag depends on the specific model being used.}
\label{tab:prompt_temp}
\end{table*}

\begin{table*}[]
\centering
\resizebox{0.75\textwidth}{!}{
\begin{tabular}{lccc|ccc}
\hline
& LLaMA 2 7B instruct & LLaMA 2 13B instruct & LLaMA 2 70B instruct& LLaMA 2 7B  & LLaMA 2 13B & LLaMA 2 70B  \\ 
\hline
Origin Acc. & 31.2\%   & 36.8\%   & 30.2\%  & 18.2\%   & 17.9\%   & 22.1\%  \\ \hline
+ $\bm{v}_{\text{honesty-fact}}$ from 7B instruct        & \textbf{36.8\%} & \textbf{40.6\%}       & \textbf{32.9\%}   & \textbf{22.6\%}   & 21.8\%   & 25.1\%        \\ 
+ $\bm{v}_{\text{honesty-fact}}$ from own instruct  & \textbf{36.8\%} & 38.1\%          & 31.3\%  & \textbf{22.6\%}   & \textbf{22.6\%}   & \textbf{25.7\%}  \\ \hline        
\end{tabular}
}
\vspace{1mm}
\caption{TruthfulQA result with $\bm{v}_{\text{honesty-fact}}$. $\bm{v}_{\text{honesty-fact}}$ of own instruct means that the vector is refined from the instruction-tuned model of the same size as the model being evaluated (e.g., LLaMA 2-13B instruct model's vector is being transplanted into LLaMA 2-13B instruct and LLaMA 2-13B).}
\label{tab:Llama 2chat_facts}
\end{table*}

\subsection{Additional Results on TruthfulQA}
\label{app:truth}

For experiments with OOD data, we  prefixed world facts generated by GPT-4 with positive/negative instructions: [\textit{Pretend you're an honest/dishonest person making statements about the world.}] to refine a new honesty vector, which is denoted as $\bm{v}_{\text{honesty-fact}}$. We used LLaMA 2-7B instruct's $\bm{v}_{\text{honesty-fact}}$ to enhance the truthfulness of LLaMA 2-13B base/instruct and LLaMA 2-70B base/instruct. Experiment results are shown in Table \ref{tab:Llama 2chat_facts}. As an OOD data source, $\bm{v}_{\text{honesty-fact}}$ can also improve the truthfulness of large models.

\begin{figure*}[]
    \vspace{-0.4cm}
    \includegraphics[width=\textwidth]{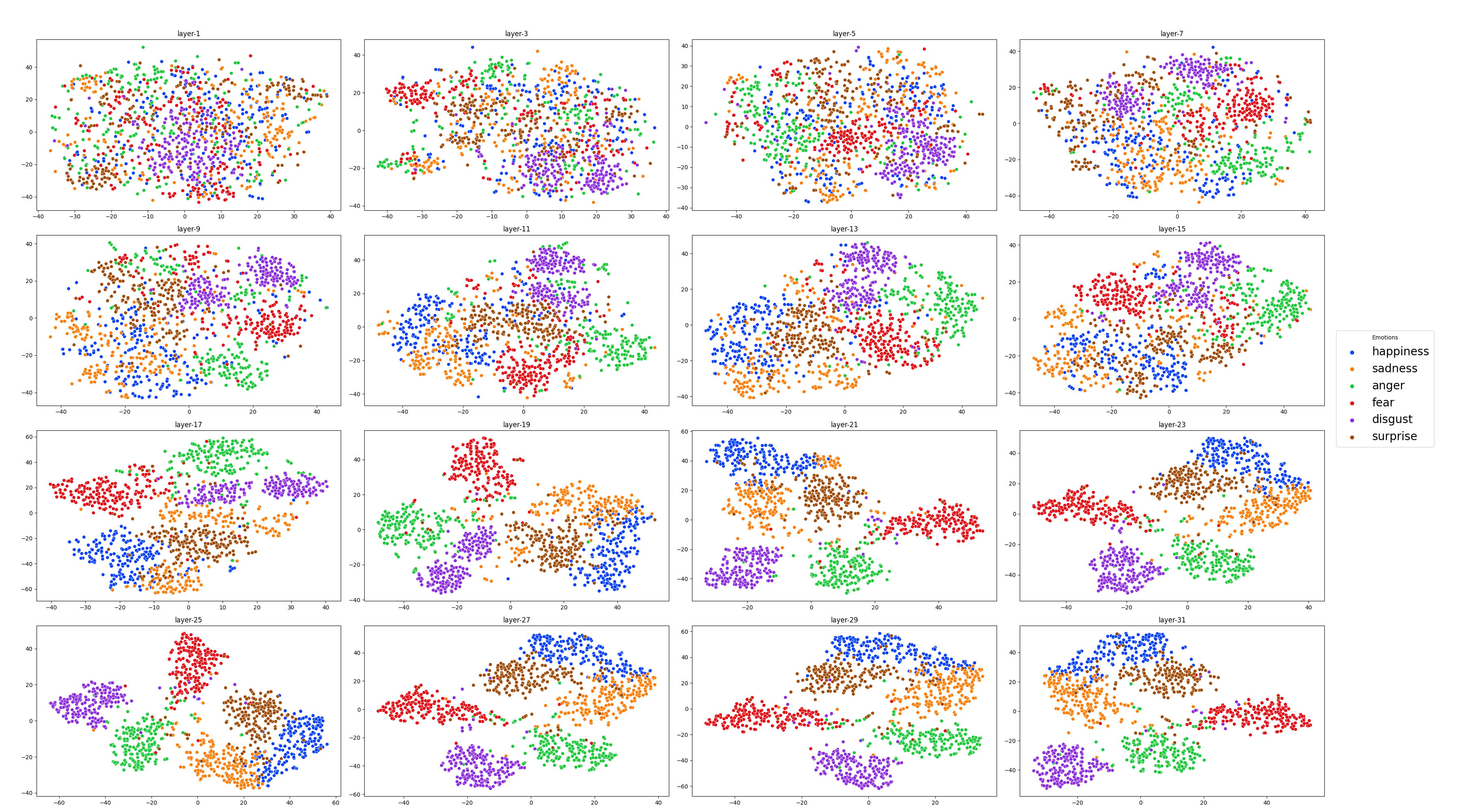}
    \caption{t-SNE visualization of hidden states of LLaMA-7B corresponding to different emotion categories.}
    \label{fig:emo_tsne}
    \vspace{-0.1cm}
\end{figure*}

\begin{figure*}[]
    \centering
    \begin{subfigure}{\textwidth}
        \centering
        \includegraphics[width=\linewidth]{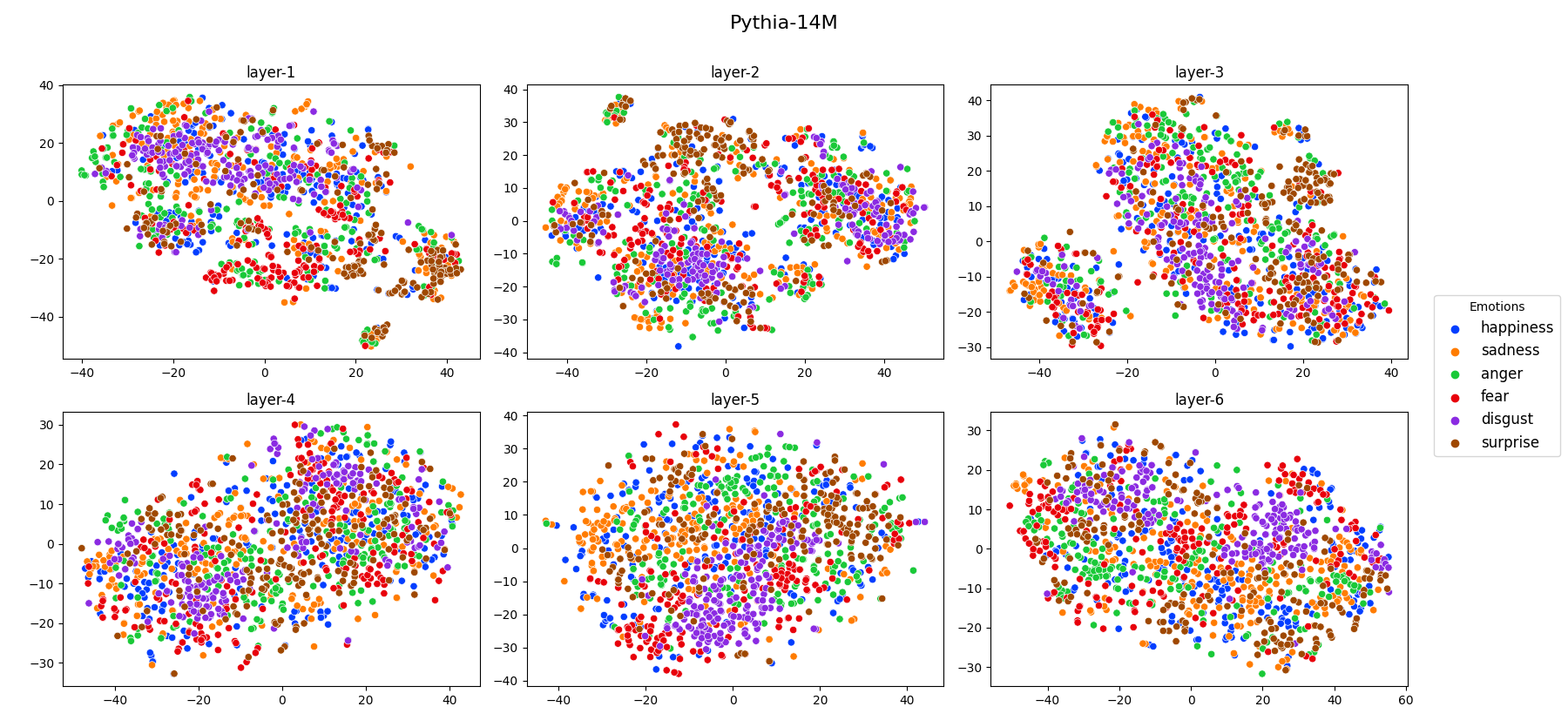}
    \end{subfigure}
    \begin{subfigure}{\textwidth}
        \centering
        \includegraphics[width=\linewidth]{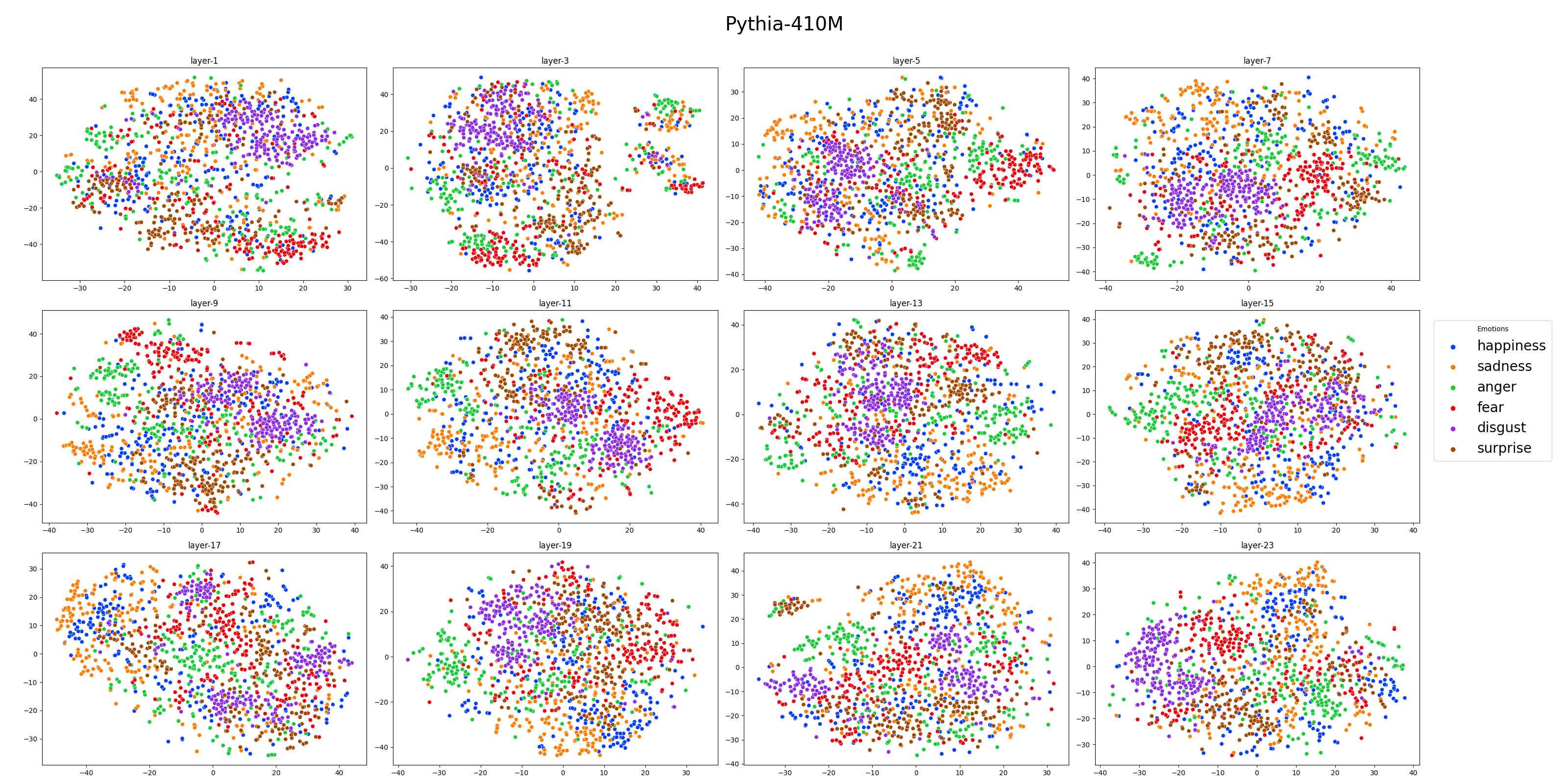}
    \end{subfigure}
    \begin{subfigure}{\textwidth}
        \centering
        \includegraphics[width=\linewidth]{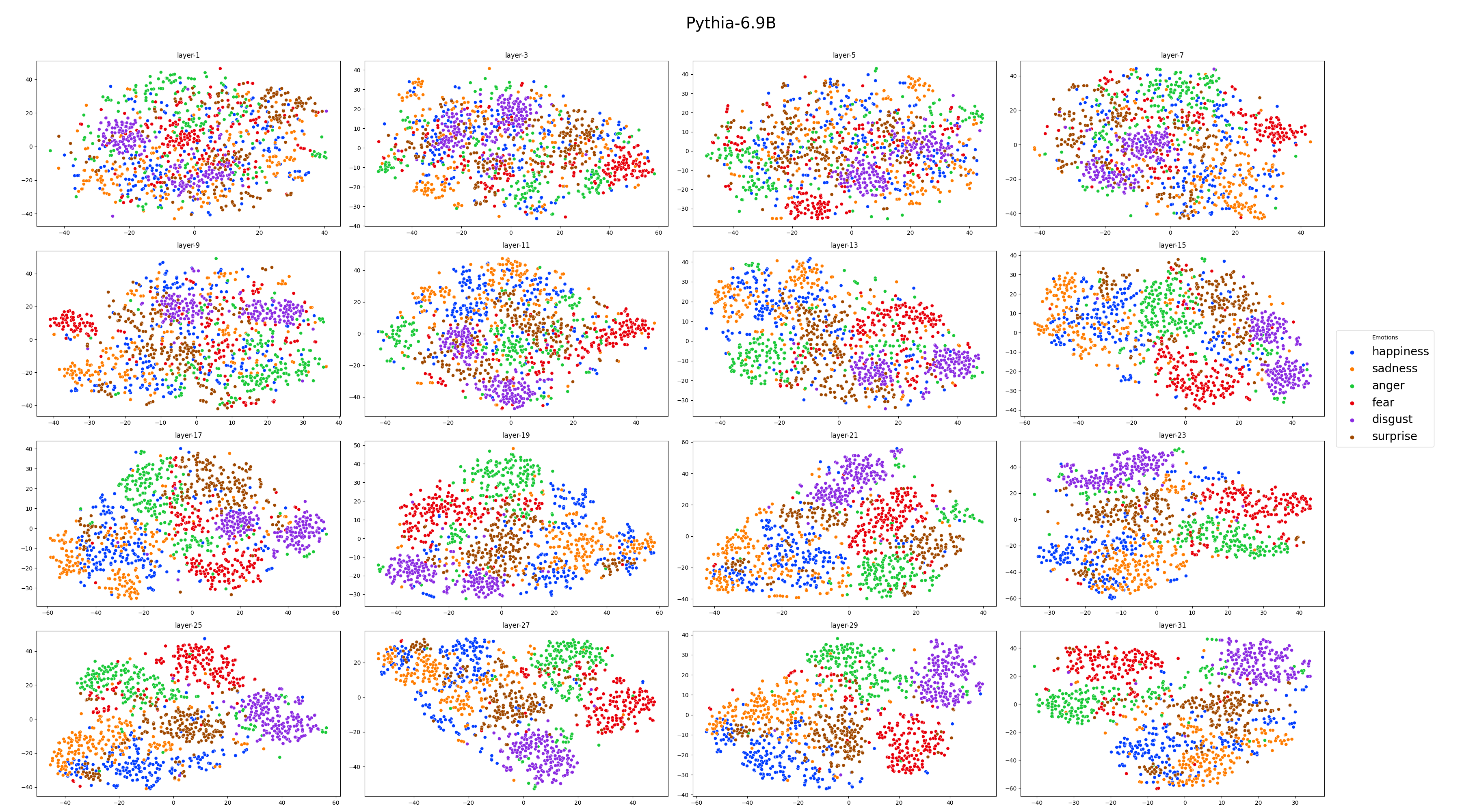}
    \end{subfigure}
    \caption{t-SNE visualization of hidden states of three Pythia models on different emotion categories. Pythia-14M cannot distinguish the features of the six emotions at all.}
    \label{fig:pythia_tsne}
\end{figure*}

\end{document}